\documentclass[11pt]{article}
\usepackage{amsfonts, amssymb, amsmath}
\usepackage{algorithm, algorithmic, amsthm}
\usepackage{graphicx}
\usepackage{color}
\usepackage{hyperref}
\usepackage[top=2.5cm, bottom=3cm, left=3cm, right=3cm]{geometry}

\usepackage{subfigure}

\usepackage{hyperref}
\usepackage{url}
\usepackage{graphicx, wrapfig}
\usepackage{amsfonts, amssymb, amsmath}
\usepackage{algorithm, algorithmic, amsthm}

\newcommand{\E}{\ensuremath{\mathbf{E}}}

\newcommand{\U}{\ensuremath{\mathbf{U}}}

\newcommand{\W}{\ensuremath{\mathbf{W}}}
\newcommand{\X}{\ensuremath{\mathbf{X}}}

\renewcommand{\b}{\ensuremath{\mathbf{b}}}

\newcommand{\f}{\ensuremath{\mathbf{f}}}

\newcommand{\h}{\ensuremath{\mathbf{h}}}

\newcommand{\q}{\ensuremath{\mathbf{q}}}

\newcommand{\w}{\ensuremath{\mathbf{w}}}
\newcommand{\x}{\ensuremath{\mathbf{x}}}
\newcommand{\y}{\ensuremath{\mathbf{y}}}
\newcommand{\z}{\ensuremath{\mathbf{z}}}

\newcommand{\calT}{\ensuremath{\mathcal{T}}}

\newcommand{\calV}{\ensuremath{\mathcal{V}}}

\newcommand{\bydef}{\stackrel{\mathrm{\scriptscriptstyle def}}{=}}

{%
\begin{list}{#1}{
\vspace{-\topsep}
\vspace{-\partopsep}
\setlength{\itemindent}{0cm}
\setlength{\rightmargin}{0cm}
\setlength{\listparindent}{0cm}
\settowidth{\labelwidth}{#1}
\setlength{\leftmargin}{\labelwidth}
\addtolength{\leftmargin}{\labelsep}
\setlength{\itemsep}{0cm}
}%
}%
{%
\end{list}
\vspace{-\topsep}
\vspace{-\partopsep}
}

{\begin{enumerate}%
}%
{\end{enumerate}}

\theoremstyle{plain}

\newtheorem*{lemma*}{Lemma}

\newtheorem*{prop*}{Proposition}

\theoremstyle{definition}

\newtheorem*{defn*}{Definition}

\newtheorem*{exmp*}{Example}

\newtheorem*{conj*}{Conjecture}

\theoremstyle{remark}

\newtheorem*{rmk*}{Remark}

\newcommand{\reason}{\textsc{Neural Reasoner }}
\newcommand{\reasonX}{\textsc{Neural Reasoner}}
\newcommand{\pr}{\textsf{Positional Reasoning} }

\newcommand{\pf}{\textsf{Path Finding} }
\newcommand{\pfX}{\textsf{Path Finding}}

\title{\sc Towards Neural Network-based Reasoning}
\author{\sf Baolin Peng$^1$\thanks{The work is done when the first author worked as intern at Noah's Ark Lab, Huawei Technologies.} \ \;Zhengdong Lu$^2$\; Hang Li$^2$ \; Kam-Fai Wong$^{1}$ \\ $^1$Dept. of Systems Engineering \& Engineering Management,\\ The Chinese University of Hong Kong\\ {\tt \{blpeng, kfwong\}@se.cuhk.edu.hk}\\ $^2$Noah's Ark Lab, Huawei Technologies\\ {\tt \{Lu.Zhengdong, HangLi.HL\}@huawei.com} }

\date{}
\begin{document}
\maketitle

\begin{abstract}
We propose \reason, a framework for neural network-based reasoning over natural language sentences. Given a question, \reason can infer over multiple supporting facts and find an answer to the question in specific forms. \reason has 1) a specific interaction-pooling mechanism, allowing it to examine multiple facts, and 2) a deep architecture, allowing it to model the complicated logical relations in reasoning tasks. Assuming no particular structure exists in the question and facts, \reason is able to accommodate different types of reasoning and different forms of language expressions. Despite the model complexity, \reason can still be trained effectively in an end-to-end manner.  Our empirical studies show that \reason can outperform existing neural reasoning systems with remarkable margins on two difficult artificial tasks (\textsf{Positional Reasoning} and \textsf{Path Finding}) proposed in \cite{DBLP:journals/corr/WestonBCM15}. For example, it improves the accuracy on \textsf{Path Finding}(10K) from $33.4\%$~\cite{memorynete2e} to over $98\%$.
\end{abstract}

\section{Introduction}
Reasoning is essential to natural language processing tasks, most obviously in examples like document summarization, question-answering, and  dialogue. Previous efforts in this direction are built on rule-based models, requiring first mapping natural languages to logic forms and then inference over them. The mapping (roughly corresponding to semantic parsing), and the inference, are by no means easy, given the variability and flexibility of natural language, the variety of the reasoning tasks, and the brittleness of a rule-based system.

Just recently, there is some new effort, mainly represented by Memory Network and its dynamic variants~\cite{memorynet,dmn}, trying to build a purely neural network-based reasoning system with fully distributed semantics that can infer over multiple facts to answer simple questions, all in natural language,  e.g., \vspace{2pt}

    \hspace{50pt}{\bf \;\;\;Fact1:} {\tt John travelled to the hallway.} \vspace{2pt}

    \hspace{50pt}{\bf \;\;\;Fact2:} {\tt Mary journeyed to the bathroom.} \vspace{2pt}

    \hspace{50pt}{\bf \;\;\;Question:} {\tt Where is Mary?} \vspace{2pt} \\
The Memory Nets perform fairly well on simple tasks like the examples above, but poorly on more complicated ones due to their simple and rigid way of modeling the dynamics of question-fact interaction and the complex process of reasoning.

In this paper we give a more systematic treatment of the problem and propose a flexible neural reasoning system, named \reasonX.  It is purely neural network based and can be trained in an end-to-end way~\cite{memorynete2e}, using only supervision from the final answer. Our contributions are mainly two-folds
\begin{itemize}
  \item we propose a novel neural reasoning system \textsc{Neural Reasoner} that can infer over multiple facts in a way insensitive to 1) the number of supporting facts, 2)the form of language, and 3) the type of reasoning;
  \item we give a particular instantiation of \reason and a multi-task training method for effectively fitting the model with relatively small amount of data, yielding significantly better results than existing neural models on two artificial reasoning task;
\end{itemize}

 %
%

\section{Overview of \reason}
\label{s:overview}
\begin{figure}[h!]
\centering
\includegraphics[width=0.8\textwidth]{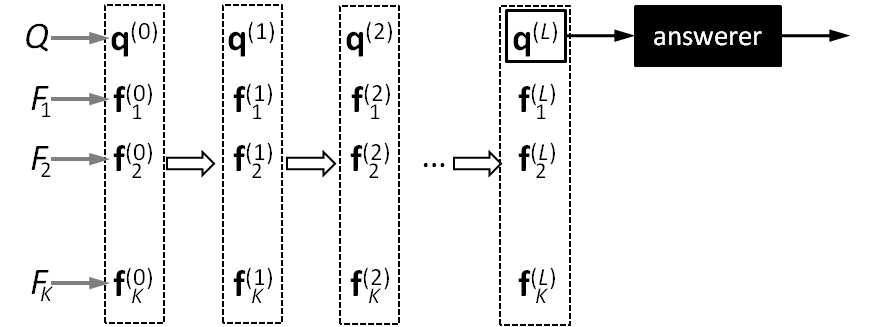}
\caption{High level system diagram of \reason.}
\label{f:HL}
\end{figure}

\reason has a layered architecture to deal with the complicated logical relations in reasoning, as illustrated in Figure~\ref{f:HL}.
 It consists of one encoding layer and multiple reasoning layers.
 The encoder layer first converts the question and facts from natural language sentences to vectorial representations. More specifically,
\[
Q \overset{\tiny \text{encode}}{-\hspace{-5pt}-\hspace{-5pt}-\hspace{-5pt}\longrightarrow} \q^{(0)}, \;\;\;
F_k \overset{\tiny \text{encode}}{-\hspace{-5pt}-\hspace{-5pt}-\hspace{-5pt}\longrightarrow}  \f_k^{(0)},\; k = 1,2,\cdots, K.
\]
where $\q^{(0)} \in \mathbb{R}^{d_Q}$ and $\f_k^{(0)} \in \mathbb{R}^{d_F}$.
With the representations obtained from the encoding layer, the reasoning layer recursively updates the representations of questions and facts, \vspace{-3pt}
\[
\{\q^{(\ell)}\;\, \f_1^{(\ell)}\;\,\cdots\;\, \f_K^{(\ell)}\} \overset{\tiny \text{reason}}{-\hspace{-5pt}-\hspace{-5pt}-\hspace{-5pt}\longrightarrow}\{\q^{(\ell+1)}\,\; \f_1^{(\ell+1)}\,\;\cdots\,\; \f_K^{(\ell+1)}\}
\]\vspace{-3pt}
through the interaction between question representation and fact representations. Intuitively, this interaction models the reasoning, including examination of the facts and comparison of the facts and the questions. Finally at layer-$L$, the resulted question representation $\q^{(L)}$ is fed to an answerer, which layer can be a classifier for choosing between a number of pre-determined classes (e.g., \{\textsf{Yes},\;\textsf{No}\}) or a text generator for create a sentence.

\newpage
We argue that \reason has the following desired properties: \vspace{-5pt}
\begin{itemize}
  \item it can handle varying number of facts, including irrelevant ones, and reach the final conclusion through repeated processing of filtering and combining; \vspace{-5pt}
  \item it makes no assumption about the form of language, as long as enough training examples are given.
\end{itemize}


\section{Model}
\begin{figure}[t!]
\centering
\includegraphics[width=0.9\textwidth]{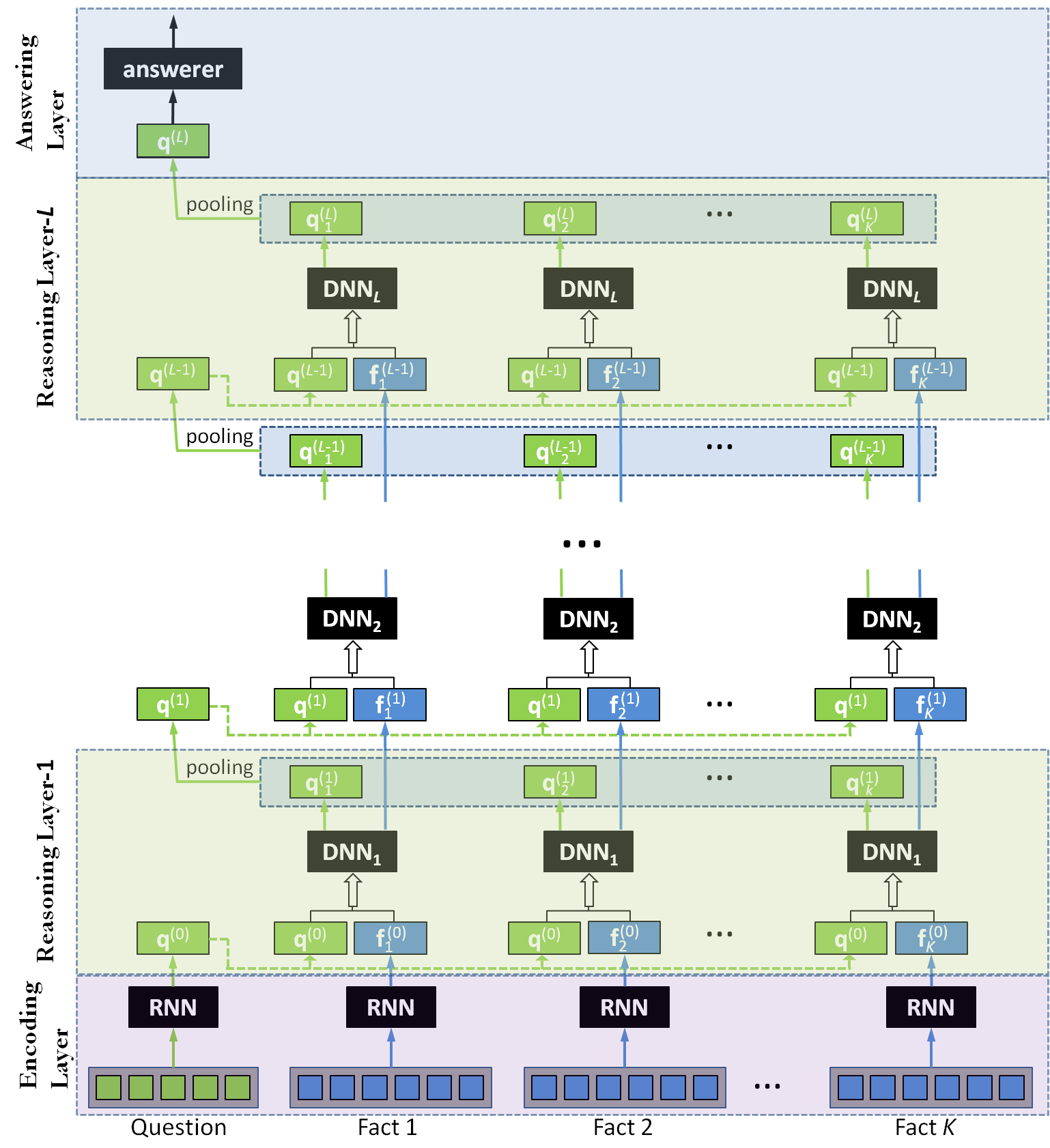}
\caption{A diagram of our implementation of \reason with $L$ reasoning layers, operating on one question and $K$ facts.}
\label{fig:NeuralReasoner}
\end{figure}

In this section we give an instantiation of \reason described in Section~\ref{s:overview}, as illustrated in Figure~\ref{fig:NeuralReasoner}. In a nutshell, question and facts, as symbol sequences, are first converted to vectorial representations in the encoding layer via recurrent neural networks (RNNs). The vectorial representations are then fed to the reasoning layers, where the question and the facts get updated through an nonlinear transformation jointly controlled by deep neural networks (DNNs)and pooling. Finally at the answering layer, the resulted question representation is used to generate the final answer to the question. More specifically 
\begin{itemize}
  \item in the encoding layer (Layer-$0$) we use recurrent neural networks (RNNs) to convert question and facts to their vectorial representations, which are then forwarded to the first reasoning layer;
  \item in each reasoning layer (i.e., Layer-$\ell$ with $1\leq \ell \leq L-1$), we use a deep neural network (denoted as \textsf{DNN}$_\ell$)  to model the pairwise interaction between question representation $\q^{(\ell-1)}$ and each fact representation $\f^{(\ell-1)}_k$  from the previous layer, which yields updated fact representation $\f^{(\ell)}_k$ and updated (fact-dependent) question representation $\q^{(\ell)}_k$;
  \item we then fuse the individual updated fact representations $\{\q^{(\ell)}_1, \q^{(\ell)}_2,\cdots, \q^{(\ell)}_K\}$ for the global updated representation $\q^{(\ell)}$ through a pooling operation (see Section \ref{s:reason} for more details)
  \item finally in Layer-$L$, the interaction net (\textsf{DNN}$_L$) returns only question update, which, after summarization by the pooling operation, will serve as input to the Answering Layer.
\end{itemize}
In the rest of this section, we will give details of different components of the model.
\subsection{Encoding Layer}

The encoding layer is designed to find semantic representations of question and facts.
Suppose that we are given a fact or a question as word sequence $\{x_1,\cdots,x_T\}$, the encoding module summarizes the word sequence with a vector with fixed length. We have different modeling choices for this purpose, e.g., CNN~\cite{NIPS14} and RNN~\cite{googleS2S}, while in this paper we use GRU~\cite{ChoEMNLP}, a variant of RNN, as the encoding module. GRU is shown to be able to alleviate the gradient vanishing issue of RNN and have similar performance to the more complicated LSTM \cite{empiricalComparison}.

\begin{figure}[h!]
\centering
       \includegraphics[width=0.82\linewidth]{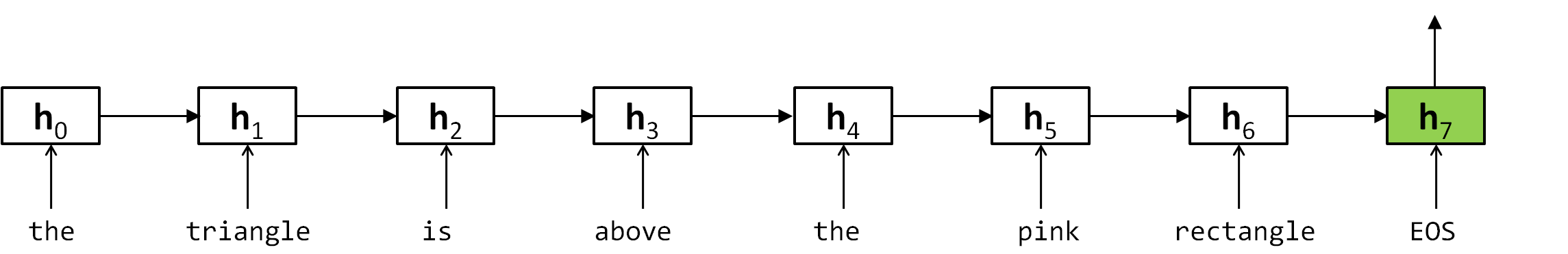}
\caption{The RNN encoder. The last state is used to summarize the word sequence.}\label{fig:encoding}
\end{figure}

As shown in Figure \ref{fig:encoding}, GRU takes as input a sequence of word vectors (for either question or facts)
\begin{eqnarray}\label{eq1}
\X = \{\x_1,\cdots, \x_T\}, \  \x_i \in \mathbb{R}^{|\calV|}
\end{eqnarray}
\noindent where $|\calV|$ stands for the size of vocabulary for input sentences. Detailed forward computations are as follows:
\begin{eqnarray}\label{eq1}
\z_t &=& \sigma(\W_\textsf{xz} \E \x_t + \W_\textsf{hz} \h_{t-1}) \\
\mathbf{r}_t &=& \sigma(\W_\textsf{xr} \E \x_t + \W_\textsf{hr} \h_{t-1})  \\
\widehat{\h}_t  &=& \tanh(\W_\textsf{xh} \E \x_t + \U_\textsf{hh} (\mathbf{r}_t \odot \h_{t-1}))\\
\h_t &=& (1 - \z_t) \odot \h_{t-1} + \z_t \odot \widehat{\h}_t
\end{eqnarray}
\noindent where $\E \in \mathbb{R}^{m \times k}$ is the word embedding  and $\W_{xz}, \W_{xr}, \W_{xh}, \W_{hz}, \W_{hr}, \U_{hh}$ are weight matrices. We take the last hidden state $\h_t$ as the representation of the word sequence.

\subsection{Reasoning Layers} \label{s:reason}
The modules in the reasoning layers include those for question-fact interaction, pooling.

\subsubsection{Question-Fact Interaction}

%

On reasoning layer $\ell$, the $k^{th}$ interaction  is between $\q^{(\ell-1)}$ and $\f_k^{(\ell-1)}$, resulting in updated representations $\q^{(\ell)}_k$ and $\f^{(\ell)}_k$
\begin{eqnarray}\label{dnnl}
[\q_k^{(\ell)}, \f^{(\ell)}_k] \bydef g_{\textsf{DNN}_\ell}( [(\q^{(\ell-1)})^\top, {\f_k^{(\ell-1)}}^\top]^\top; \Theta_\ell),
\end{eqnarray}
with $\Theta_\ell$ being the parameters. In general, $\q^{(\ell)}_k$ and $\f^{(\ell)}_k$ can be of different dimensionality as those of the previous layers.
In the simplest case with a single layer in \textsf{DNN}$_\ell$, we have
\begin{eqnarray}\label{eq1}
\q_k^{(\ell)} \bydef \sigma(\W_\ell^\top[(\q^{(\ell-1)})^\top, {\f_k^{(\ell-1)}}^\top] + \b_\ell),
\end{eqnarray}
where $ \sigma(\cdot)$ stands for the nonlinear activation function.

Roughly speaking, $\q^{(\ell)}_k$ contains the update of the system's understanding on answering the question after its interaction with fact $K$, while $\f^{(\ell)}_k$ records the change of the $K^{th}$ fact. Therefore, $\{(\q^{(\ell)}_k$ ,$\f^{(\ell)}_k)\}$ constitute the ``state" of the reasoning process.


\subsubsection{Pooling}
Pooling aims to fuse the understanding of the question right after its interaction with all the facts to form the current status of the question, through which we can enable the comparison between different facts. There are several strategies for this pooling
\begin{itemize}
  \item {\bf Average/Max Pooling:} To obtain the $n^{th}$ element in $\q^{(\ell)}$, we can take the average or the maximum of the elements at the same location from $\{\q^{(\ell)}_1,\cdots \q^{(\ell)}_K\}$. For example, with max-pooling, we have
      \[
      \q^{(\ell)}(d) = \max(\{\q_1^{(\ell)}(d),\q_2^{(\ell)}(d),\cdots, \q_K^{(\ell)}(d)\}), \;\;d = 1,2,\cdots, D_\ell
      \]
      where $\q^{(\ell)}(d)$ stands for the $d^{th}$ element of vector $\q^{(\ell)}$.
      Clearly this kind of pooling is the simplest, without any associated parameters;
  \item {\bf Gating:} We can have an extra gating network $g^{(\ell)}(\cdot)$ to determine the certainty of the features in $\q^{(\ell)}_k$ based on $\{ \q^{(\ell-1)}, \f^{(\ell-1)}_k \}$ (the input for getting $\q^{(\ell)}_k$). The output $g^{(\ell)}(\q^{(\ell-1)}, \f^{(\ell-1)}_k )$ has the same dimension as $\q^{(\ell)}_k$, whose $n^{th}$ element, after normalization, can be used as weight for the corresponding element in $\q^{(\ell)}_k$ in obtaining $\q^{(\ell)}$.
  \item {\bf Model-based:} In the case of temporal-reasoning, there is crucial information in the sequential order of the facts. To account for this temporal structure, we can use a CNN or RNN to combine the information in $\{\q^{(\ell)}_1,\cdots \q^{(\ell)}_K\}$.
\end{itemize}
At layer-$L$, the query representation $\q^{(L)}$ after the pooling will serve as the features for the final decision.

\subsection{Answering Layer}

For simplicity, we focus on the reasoning tasks which can be formulated as classification with predetermined classes. More specifically, we apply \reason to deal with the following two types of questions
\begin{itemize}
  \item {\bf Type I:}  General questions, i.e., questions with Yes-No answer;
  \item {\bf Type II: } Special questions with a small set of candidate answers.
\end{itemize}
At reasoning Layer-$L$, it performs pooling over the intermediate results to select important information for further uses.
\begin{eqnarray}\label{eq1}
\q &=& \textsf{pool}(\{ \q^{(L)}_1, \q^{(L)}_2, \cdots, \q^{(L)}_K\}) \\
\y &=& \textsf{softmax}(\W^\top_\textsf{softmax} \q^{(L)})
\end{eqnarray}
After reaching the last reasoning step, in this paper we take two steps, $Q^2$ is sent to a standard softmax layer to generate an answer which is formulated as a classification problem.

There is another type of prediction as classification where the \emph{effective} classes dynamically change with instances, e.g., the \textsf{Single-Supporting-Fact} task in \cite{memorynet}.  Those tasks cannot be directly solved with \textsc{Neural Reasoner}. One simple way to  circumvent this is to define the following score function
\[
\textsf{score}_z = g_{\textsf{match}} (\q^{(L)},\w_z;\theta)
\]
where $ g_{\textsf{match}}$  is a function (e.g., a DNN) parameterized with $\theta$, and  $\w_z$ is the embedding for class $z$, with $z$ being dynamically determined for the task.

\subsection{Training} \label{s:training}
The training of model tunes the parameters in $\{\textsf{RNN}_0, \textsf{DNN}_1,\cdots, \textsf{DNN}_L \}$ and those in the softmax classifier.  Similar to \cite{memorynete2e}, we perform end-to-end training, taking the final answer as the only supervision. More specifically,
We use the cross entropy for the cost of classification
\[
E_\textsf{reasoning} = \sum_{n\in \calT} D_{\textsf{CE}}(p(\y|r_n)||\y_n)
\]
where $n$ indexes the instances in the training set $\calT$, and $r_n = \{Q_n, F_{n,1},\cdots, F_{n, K_n}\}$ stands for question and facts for the $n^{th}$ instance.

Our end-to-end training is the same as~\cite{memorynete2e}, while the training in~\cite{memorynet}and~\cite{dmn} use the step-by-step labels on the supporting facts for each instance (see Table~\ref{tab:sampels} for examples) in addition to the answer. As described in~\cite{memorynete2e}, those extra labels brings much stronger supervision just the answer in the end-to-end learning setting, and typically yield significantly better result on relatively complicated tasks.

\section{Auxiliary Training for Question/Fact Representation} \label{s:multitask}



We use auxiliary training to facilitate the learning of representations of question and facts. Basically, in addition to using the learned representations of question and facts in the reasoning process, we also use those representations to reconstruct the original questions or their more abstract forms with variables (elaborated later in Section~\ref{s:variables}).

In the auxiliary training, we intend to achieve the following two goals \vspace{-5pt}
\begin{itemize}
  \item to compensate the lack of supervision in the learning task.  In our experiments, the supervision can be fairly weak since for each instance it is merely a classification with no more than 12 classes, while the number of instances are 1K to 10K. \vspace{-5pt}
\item to introduce beneficial bias for the representation learning task. Since the network is a complicated nonlinear function,  the back-propagation from the answering layer to the encoding layer can easily fail to learn well.
\end{itemize}

\begin{figure}[h!]
\begin{center}
            \includegraphics[width=0.42\textwidth]{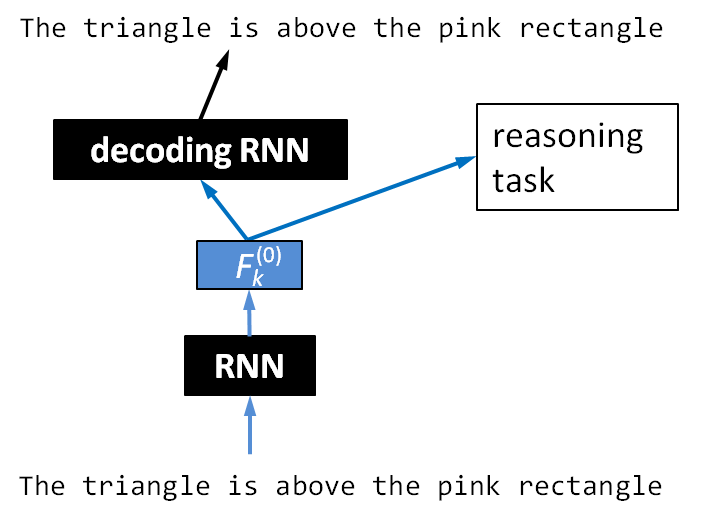}\hspace{40pt}
            \includegraphics[width=0.42\textwidth]{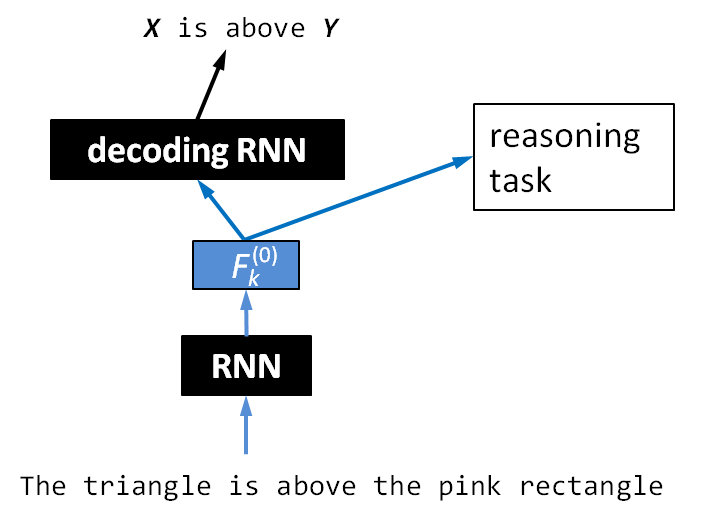}\\
\textsf{\small (a) reconstructing the original sentence} \hspace{50pt}  \textsf{\small (b) producing an abstract form with variables}
      \caption{Auxiliary training for question representation. The training for fact representation is identical and therefore omitted.}
    \label{f:auxTask}
  \end{center}
\end{figure} \vspace{-10pt}

\subsection{Multi-task Learning Setting}
As illustrated in Figure~\ref{f:auxTask}, we take the simplest way to fuse the auxiliary tasks (recovering) with the main task (reasoning) through linearly combining their costs with trade-off parameter $\alpha$
\begin{eqnarray}\label{eq1}
E = \alpha E_\textsf{recovering} + (1-\alpha) E_\textsf{reasoning}
\end{eqnarray}
\noindent where$E_\textsf{reasoning}$ is the cross entropy loss describing the discrepancy of model prediction from correct answer (see Section~\ref{s:training}), and  $E_\textsf{recovering}$ is the negative log-likelihood of the sequences (question or facts) to be recovered. More specifically,
\[
E_\textsf{recovering} = \sum_{n\in \calT} \{ \sum_{k=1}^{K_n}\log p(F_{n,k}| \f_{n,k}^{(0)}) +  \log p(Q_n | \q_n^{(0)})\}
\]
where the likelihood is estimated as in the encoder-decoder framework proposed in \cite{ChoEMNLP}. On top of the encoding layer (RNN), we add another decoding layer (RNN) which is trained to sequentially predict words in the original sentence.

\subsection{Abstract Forms with Variables} \label{s:variables}
Instead of recovering the original sentence in question and facts, we also study the effect of producing a more abstract form in the auxiliary training task. More specifically, we let the decoding RNN to recover a sentence with entities replaced with variables (treated as particular symbols), e.g.,
\begin{eqnarray*}
\texttt{\small  {\color{blue}The triangle} is above  {\color{blue}the pink rectangle}} .\hspace{-10pt}
 &\overset{\tiny \text{recover}}{-\hspace{-5pt}-\hspace{-5pt}-\hspace{-5pt}\longrightarrow} \hspace{-10pt} &
x\;  \texttt{\small is above } y. \\
\texttt{\small   {\color{blue}The blue square} is to the left of  {\color{blue}the triangle}}. \hspace{-10pt}
 &\overset{\tiny \text{recover}}{-\hspace{-5pt}-\hspace{-5pt}-\hspace{-5pt}\longrightarrow} \hspace{-10pt}  &
z\;  \texttt{\small is to the left of } x. \\
\texttt{\small
Is  {\color{blue}the pink rectangle} to the right of  {\color{blue}the square}?}\hspace{-10pt}
 &\overset{\tiny \text{recover}}{-\hspace{-5pt}-\hspace{-5pt}-\hspace{-5pt}\longrightarrow} \hspace{-10pt} &
\texttt{\small Is $y$ to the right of the $z$ ?}
\end{eqnarray*}

Through this, we intend to teach the system a more abstract way of representing sentences (both question and facts) and their interactions. More specifically,
\begin{itemize}
  \item all the entities are only meaningful only when they are compared with each other. In other words, the model (in the encoding and reasoning layers) should not consider specific entities, but their general notions.
  \item it helps the model to focus on the relations between the entities,  the commonality of different facts, and the patterns shared between different instances.
\end{itemize}


\section{Experiments}
We report our empirical study on applying \reason to the Question Answer task defined in \cite{DBLP:journals/corr/WestonBCM15}, and compare it against state-of-the-art neural models~\cite{memorynet,dmn}.
\subsection{Setup}
bAbI is a synthetic question and answering dataset. It contains 20 tasks, and each of them is composed of a set of facts, a question and followed by an answer which is mostly a single word. For most of the time, only a subset of facts are relevant to the given question. Two versions of the data are available, one has 1K training instances per task and the other has 10K instances per task, while the testing set are the same for the two versions.

We select the two most challenging tasks (among the 20 tasks in  \cite{DBLP:journals/corr/WestonBCM15} ) \pr and \pfX, to test the reasoning ability of \reasonX. \pr task tests model's spatial reasoning ability, while \pf task, first proposed in \cite{DBLP:conf/aaai/ChenM11} tests the ability to reason the correct path between objects based on natural language instructions. In Table \ref{tab:sampels}, we give an instance of each task.

%
\begin{table}[htdp]

\hspace{25pt}\textsf{Task I: path finding} \vspace{-5pt}
\begin{center}
\hspace{-12pt}
\begin{tabular}{|l|}
\hline
\tt \small
1.The office is east of the hallway. \\
\tt \small
2.The kitchen is north of the office. \\
\tt \small
3.The garden is west of the bedroom. \\
\tt \small
4.The office is west of the garden.\\
\tt \small
5.The bathroom is north of the garden. \\
\tt \small
How do you go from the kitchen to the garden? \rm  {\color{red}south, east}, relies on 2 and 4\\
\tt \small
How do you go from the office to the bathroom? \rm {\color{red}east, north}, relies on 4 and 5\\

\hline
\end{tabular}
\end{center}

\hspace{25pt}\textsf{Task II: positional reasoning} \vspace{-5pt}
\begin{center}
\begin{tabular}{|l|}
\hline
\tt \small
1.The triangle is above the pink rectangle. \\
\tt \small
2.The blue square is to the left of the triangle.  \\
\tt \small
Is the pink rectangle to the right of the blue square?  \rm {\color{red}Yes}, relies on 1 and 2\\
\tt \small
Is the blue square below the pink rectangle?	\rm {\color{red}No}, relies on 1 and 2 \\
\hline
\end{tabular}
\end{center}

\caption{Samples of the two tasks: \textsf{path finding} (upper panel) and \textsf{positional reasoning} (lower panel), with facts, questions and given answers (following each question). For each panel, we first list facts and then question that one needs to answer based on the given facts. On Task I, the answer to the first question is
$south,east$, standing for going south first and then east, obtained based on fact 2 and 4. }
\label{tab:sampels}
\end{table}%

\subsection{Implementation Details}
In our experiments, we actually used a simplified version of \reason. In the version
\begin{itemize}
  \item we choose to keep the representation un-updated on each layer, e.g.,
  \[
 F_k \overset{\tiny \text{encode}}{-\hspace{-5pt}-\hspace{-5pt}-\hspace{-5pt}\longrightarrow}  \f^{(0)}_k = \f^{(1)}_k=\cdots = \f^{(L-1)}_k, \; k = 1,2,\cdots, K.
  \]
  This choice pushes the update $\q^{(\ell)}_k$ (and its summarization $\q^{(\ell)}$) to record all the information in the interaction between facts and question. 
  \item we use only two layers, i.e., $L=2$, for the relatively simple task in the experiments.
\end{itemize}

Our model was trained with the standard back-propagation (BP) aiming to maximize the likelihood of correct answers. All the parameters including the word-embeddings were initialized by randomly sampling from a uniform distribution [-0.1, 0.1]. No momentum and weight decay was used. We trained all the tasks for 200 epochs with stochastic gradient descent and the gradients which had $\ell_2$ norm larger than 40 were clipped, learning rate being controlled by AdaDelta \cite{adadelta}. For multi-task learning, different mixture ratios were tried, from 0.1 to 0.9.

\subsection{\reason vs. Competitor Models}
We compare \reason with the following three neural reasoning models: 1)Memory Network, including the one with step-by-step supervision~\cite{memorynet}(denoted as \textsc{Memory Net-Step}) and the end-to-end version~\cite{memorynete2e} (denoted as \textsc{Memory Net-N2N}), and 2) \textsc{Dynamic Memory Network}, proposed in \cite{dmn}, also with step-by-step supervision. 
In Table~\ref{t:results}, we report the performance of a particular case of \reason with 1) two reasoning layers, 2) 2-layer DNNs as the interaction modules in each reasoning layer, and 3) auxiliary task of recovering the original question and facts.
The results are compared against three neural competitors. We have the following observations.
\begin{itemize}
  \item The proposed \reason performs significantly better than Memory Net-N2N, especially with more training data.
  \item Although not a fair comparison to our model,  \reason is actually better than \textsc{Memory Net-N2N} and \textsc{Dynamic Memory Net} on \textsf{Positional Reasoning} (1K) \& (10K) as well as \textsf{Path Finding} (10K), with about 20\% margin on both tasks with 10K training instances.
\end{itemize}

\begin{table}[htdp] \footnotesize
\begin{center}
\begin{tabular}{|l|c|c|}
\hline
   &\textsf{\small Posi. Reason.} (1K) & \textsf{\small Posi. Reason.} (10K)\\
         \hline
         \hline
        {\bf Step-by-step Supervision} && \\
        \hspace{50pt}\textsc{Memory Net-step}  & 65.0\% & 75.4\% \\
       \hspace{50pt}\textsc{Dynamic Memory Net}  & 59.6\%&-  \\
        \hline
        \hline
                {\bf End-to-End} && \\
        \hspace{50pt}\textsc{Memory Net-N2N} & 59.6\% &60.3\%\\
        \hspace{50pt}{\small \reason}  &  66.4\% & 97.9\%\\
        \hline
\end{tabular}
\end{center}

\begin{center}
\begin{tabular}{|l|c|c|}
\hline
   &\textsf{\small Path Finding} \;(1K) & \textsf{\small Path Finding}
   \; (10K)\\
         \hline
         \hline
        {\bf Step-by-step Supervision} && \\
        \hspace{50pt}\textsc{Memory Net-step}  & 36.0\% & 68.1\% \\
       \hspace{50pt}\textsc{Dynamic Memory Net}  & 34.5\%&-  \\
        \hline
        \hline
                {\bf End-to-End} && \\
        \hspace{50pt}\textsc{Memory Net-N2N} & 17.2\% &33.4\%\\
        \hspace{50pt}{\small \reason}  &  17.3\% & 87.0\%\\
        \hline
\end{tabular}
\end{center}
\caption{Results on two reasoning tasks. The results of \textsc{Memory Net-step}, \textsc{Memory Net-N2N}, and \textsc{Dynamic Memory Net} are taken respectively from \cite{memorynet},\cite{memorynete2e} and \cite{dmn}.}
\label{t:results}
\end{table}%

Please note that the results of \reason reported in Table~\ref{t:results} are not based on architectures specifically tuned for the tasks. As a matter of fact, with more complicated models (more reasoning layers and deeper interaction modules), we can achieve even better results on large datasets (e.g., over $98\%$ accuracy on \textsf{Path Finding} with 10K instances). We will however leave the discussion on different architectural variants to the next section.

\subsection{Architectural Variations}
This section is devoted to the study of architectural variants of \reasonX. More specifically, we consider the variations in 1)the number of reasoning layers, 2) the depth of the interaction DNN, and 3) the auxiliary tasks, with results summarized by Table~\ref{t:results_variations}. We have the following observations:
\begin{itemize}
\item  Auxiliary tasks are essential to the efficacy of \reasonX, without which the performances of \reason drop dramatically. The reason, as we conjecture in Section~\ref{s:multitask}, is that the reasoning task alone cannot give enough supervision for learning accurate word vectors and parameters of the RNN encoder. We note that \reason can still outperform Memory Net (N2N) with 10K data on both tasks.
  \item \reason with shallow architectures, more specifically two reasoning layers and 1-layer DNN, apparently can benefit from the auxiliary learning of recovering abstract forms on small datasets (1K on both tasks). However, with deeper architectures or more training data, the improvement over that of recovering original sentences become smaller, despite the extra information it utilizes.
  \item When larger training datasets are available, \reason appears to prefer relatively deeper architectures. More importantly, although both tasks require two reasoning steps, the performance does not deteriorate with three reasoning layers. On both tasks, with 10K training instances, \reason with three reasoning layers and 3-layer DNN can achieve over $98\%$ accuracy.

\end{itemize}

\begin{table}[t!] \footnotesize
\begin{center}
\begin{tabular}{|l|c|c|}
\hline
   &\textsf{\small Posi. Reason.} (1K) & \textsf{\small Posi. Reason.} (10K)\\
         \hline
  {\bf  No auxiliary task} && \\
     \hspace{65pt}2-layer reasoning, 1-layer DNN & 60.2\%& 72.1\% \\
      \hspace{65pt}2-layer reasoning, 2-layer DNN &   59.6\%  & 69.3\%\\
\hspace{65pt}3-layer reasoning, 3-layer DNN &   58.7\%  & 59.7\%\\
        \hline
        {\bf  Auxiliary task: Original } && \\
     \hspace{65pt}2-layer reasoning, 1-layer DNN & 63.1\%&93.8\% \\
      \hspace{65pt}2-layer reasoning, 2-layer DNN  &   66.4\%  & 97.9\%\\
      \hspace{65pt}3-layer reasoning, 3-layer DNN &   69.4\%   & 99.1\% \\
        \hline \hline
         {\bf  Auxiliary task: Abstract } && \\
     \hspace{65pt}2-layer reasoning, 1-layer DNN& 70.9\%&95.2\% \\
     \hspace{65pt}2-layer reasoning, 2-layer DNN &   66.6\% & 95.6\%\\
        \hspace{65pt}3-layer reasoning, 3-layer DNN &   68.3\%  & 97.4\%\\
        \hline
\end{tabular}
\end{center}

\begin{center}
\begin{tabular}{|l|c|c|}
\hline
   &\textsf{\small Path Finding} \;(1K) & \textsf{\small Path Finding}
   \; (10K)\\
         \hline
  {\bf  No auxiliary task} && \\
     \hspace{65pt}2-layer reasoning, 1-layer DNN & 13.6\%& 52.2\% \\
      \hspace{65pt}2-layer reasoning, 2-layer DNN &   12.3\%  & 54.2\%\\
\hspace{65pt}3-layer reasoning, 3-layer DNN &   13.1\%  & 51.7\%\\
        \hline
        {\bf  Auxiliary task: Original } && \\
     \hspace{65pt}2-layer reasoning, 1-layer DNN & 14.1\%&57.0\% \\
      \hspace{65pt}2-layer reasoning, 2-layer DNN  &   17.3\%  & 87.0\%\\
      \hspace{65pt}3-layer reasoning, 3-layer DNN &   14.0\%  & 98.4\%\\
        \hline \hline
         {\bf  Auxiliary task: Abstract } && \\
     \hspace{65pt}2-layer reasoning, 1-layer DNN& 18.1\%&55.8\% \\
     \hspace{65pt}2-layer reasoning, 2-layer DNN &   15.4\% & 87.8\%\\
        \hspace{65pt}3-layer reasoning, 3-layer DNN &   11.3\%  & 98.6\%\\
        \hline
\end{tabular}
\end{center}

\caption{Results on two reasoning tasks yielded by \reason with different architectural variations.}
\label{t:results_variations}
\end{table}%

\section{Conclusion and Future Work}
We have proposed \reasonX, a framework for neural network-based reasoning over natural language sentences. \reason is flexible, powerful, and language indepedent. Our empirical studies show that \reason can dramatically improve upon existing neural reasoning systems on two difficult artificial tasks proposed in \cite{memorynet}. For future work, we will explore 1) tasks with higher difficulty and reasoning depth, e.g., tasks which require a large number of supporting facts and facts with complex intrinsic structures, 2)
the common structure in different but similar reasoning tasks (e.g., multiple tasks all with general questions), and 3)
automatic selection of the reasoning architecture, for example, determining when to stop the reasoning based on the data.

%
%
%
%
%

\bibliographystyle{abbrv}
\small{\bibliography{nips2015}}
\end{document}